%% file: main.tex
\documentclass[final]{cvpr}

\usepackage{times}
\usepackage{epsfig}
\usepackage{graphicx}
\usepackage{amsmath}
\usepackage{amssymb}
\usepackage{dsfont}
\usepackage[utf8]{inputenc}
\usepackage{multirow}
\usepackage[usenames,dvipsnames,svgnames,table]{xcolor}
\usepackage[ruled,vlined,linesnumbered]{algorithm2e} 
\usepackage{tablefootnote} 
\graphicspath{{./}{./images/}}
\usepackage{subcaption}
\usepackage[pagebackref=true,breaklinks=true,colorlinks,bookmarks=false]{hyperref}

\def\cvprPaperID{4070} 
\def\confYear{CVPR 2021}

\makeatletter
\DeclareRobustCommand\onedot{\futurelet\@let@token\@onedot}
\def\@onedot{\ifx\@let@token.\else.\null\fi\xspace}

\def\eg{\emph{e.g}\onedot} 
\def\ie{{i.e}\onedot}

\def\etal{{et al}\onedot}
\makeatother

\def\Vec#1{{\boldsymbol{#1}}}

\newcommand\norm[1]{\left\lVert#1\right\rVert} 

\DeclareMathOperator*{\argmax}{arg\,max}

\title{All Labels Are Not Created Equal: Enhancing Semi-supervision via Label Grouping and Co-training}
\author{Islam Nassar\textsuperscript{1}\thanks{corresponding author: islam.nassar@monash.edu}, Samitha Herath\textsuperscript{1}, Ehsan Abbasnejad\textsuperscript{2}, Wray Buntine\textsuperscript{1}, Gholamreza Haffari\textsuperscript{1}\\
\textsuperscript{1} Dept of Data Science and AI, Faculty of IT, Monash University, Australia\\
\textsuperscript{2} Australian Institute for Machine Learning, University of Adelaide, Australia

}
\date{October 2020}

\begin{document}
\maketitle    

\begin{abstract}
Pseudo-labeling is a key component in semi-supervised learning (SSL). It relies on iteratively using the model to generate artificial labels for the unlabeled data to train against. A common property among its various methods is that they only rely on the model's prediction to make labeling decisions without considering any prior knowledge about the visual similarity among the classes. In this paper, we demonstrate that this degrades the quality of pseudo-labeling as it poorly represents visually similar classes in the pool of pseudo-labeled data. We propose SemCo, a method which leverages label semantics and co-training to address this problem. We train two classifiers with two different views of the class labels: one classifier uses the one-hot view of the labels and disregards any potential similarity among the classes, while the other uses a distributed view of the labels and groups potentially similar classes together. We then co-train the two classifiers to learn based on their disagreements. We show that our method achieves state-of-the-art performance across various SSL tasks including 5.6\% accuracy improvement on Mini-ImageNet dataset with 1000 labeled examples. We also show that our method requires smaller batch size and fewer training iterations to reach its best performance. We make our code available at~\url{https://github.com/islam-nassar/semco}.

\end{abstract}

\input{1Introduction}
\input{2Method}
\input{3Experiments}

\input{4RelatedWork}
\input{5Conclusion}

{\small
\bibliographystyle{ieee_fullname}
\bibliography{references.bib}
}
\newpage

\input{6Supplementary}
\end{document}

%% file: 1Introduction.tex
\section{Introduction}
\label{introduction}
Deep neural models require large amounts of labeled data to achieve their high performance. This quickly becomes prohibitive and non-scalable especially when labeling data is expensive and/or non practical. Semi-supervised learning (SSL)~\cite{chapelle2009semi, van2020survey} has hence emerged to explore a diverse set of methods which aim to leverage unlabeled data to enable learning from a smaller set of labeled data.

In the context of image classification, recent methods use unlabeled data to guide learning in different ways. Some methods primarily focus on consistency regularization~\cite{Sajjadi_NIPS16, Laine_temporal16}, where the model is enforced to produce consistent predictions for different perturbed versions of the same unlabeled input image. While others focus on pseudo-labeling~\cite{arazo_pseudo, lee2013_pseudo, label_prop}, where the model is used to produce artificial labels for the unlabeled data that are then used to further train the model. Evidently, combining the two approaches has  shown the state-of-the-art results on various image classification tasks~\cite{Sohn_fixmatch20}.

\begin{figure}[!t]
 \centering
  \includegraphics[width=0.43\textwidth]{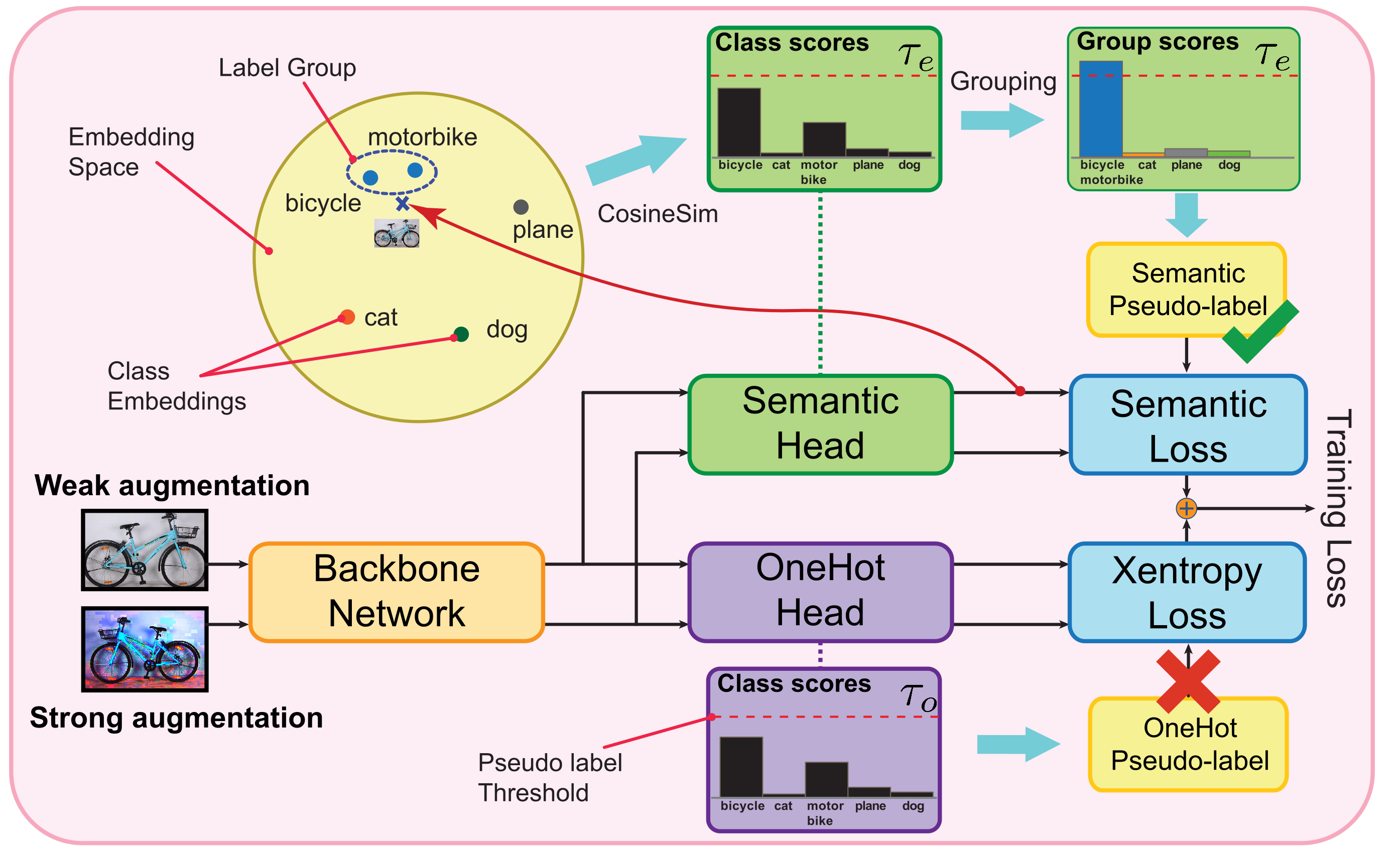}
 \caption{A conceptual diagram of our co-training solution}
 \vspace{-5mm}
 \label{fig:high_level_diagram}
\end{figure}

When it comes to pseudo-labeling, a common problem which hinders the SSL performance is the so-called confirmation bias~\cite{mean_teacher}. This takes place when the model reassures its wrong predictions by retraining on them, leading to an accumulation of the error from which the model can not recover. To mitigate this behaviour, some methods use a warm-up phase until the model becomes more reliable~\cite{label_prop, mean_teacher}, or limit the number of pseudo-labeled samples in each mini-batch~\cite{arazo_pseudo}. Other strategies include using a confidence threshold whereby a sample is only considered for pseudo-labeling if the model is highly confident about its prediction~\cite{Sohn_fixmatch20, lee2013_pseudo}. One property shared by all these methods, however, is that they only rely on the model's output and disregard any prior knowledge about potential similarities among the classes. As we show in Section~\ref{pseudo_background}, visually similar classes are expected to confuse the model and therefore get poorly represented in the pseudo-labeled data pool. This fact is even more exacerbated in confidence-based methods~\cite{lee2013_pseudo, Sohn_fixmatch20} as it leads to discarding most of the visually similar samples simply because the model is rarely confident about their predictions. We show that this leads to a class imbalance in the pseudo-labeled pool, and thereby, misguides the training.

In this paper, we demonstrate that by exploiting class labels semantics, we can account for such similarity among the classes. We draw inspiration from few-shot learning methods ~\cite{frome2013devise, ye2018learning_embedding} where we use distributed embeddings to represent class labels. We present two methods to generate label embeddings in a way which encodes a weak prior on the visual similarity among the classes. One such method is based on knowledge graph embeddings~\cite{speer2016_conceptnet}, while another is based on visual attributes annotation~\cite{cub200}.  Having such embeddings provides basis to group the class labels into visually similar concepts and allow considering such grouping while making pseudo-labeling decisions. 

The benefit of using label embeddings goes beyond label grouping. Earlier work~\cite{frome2013devise} has shown that using embeddings as training targets (as opposed to one-hot labels) allows the model to map the image features to a more meaningful semantic space, and thereby, enables few shot transfer. In our work, we  leverage this idea in a  co-training~\cite{blum1998_co_training} style approach to improve SSL performance. We propose to train two classifiers with the two different views of the class labels, i.e. one-hot and distributed. One of the classifiers makes use of the label grouping during pseudo-labelling, while the other does not. We then allow the two classifiers to learn from their disagreements via a shared consistency regularization loss on the unlabeled data.  

We show that our method achieves new state-of-the-art results across five different datasets, while using smaller batch size  with fewer training iterations.
%
To summarize, our contributions are:
\begin{enumerate}
\itemsep0em 
    \item We propose an approach which leverages the semantic similarity among the classes to improve pseudo-labeling quality by addressing the confusion events. 
    \item We present a co-training-based SSL method which involves two classifiers co-operating via pseudo-labels obtained using their  different views of the class label. 
    \item We show our approach outperforms the state-of-the-art in SSL by a large margin on 5 different datasets including 5.6\% on Mini-Imagenet with 1000 labeled point, i.e. 10 labels per class.
\end{enumerate}

%% file: 2Method.tex
\section{Background}
\label{sec:background}
We are interested in a $K$-way semi-supervised image classification problem, where we train a model using batches of both labelled and unlabelled examples.
Specifically, each batch comprises labeled examples, $\mathcal{X} = \{(\Vec{x}_i, \Vec{y}_i)\}_{i=1}^n$ and unlabeled examples, $\mathcal{U} = \{\Vec{u}_j\}_{j=1}^{\mu \cdot n}$, where the scalar $\mu$ denotes the ratio between the number of unlabeled and labeled examples in a given batch, and $\Vec{y}_i$ denotes the one-hot representation of the label. But before we introduce our approach, we begin by reviewing three key concepts underpinning our method.  

\smallskip
\noindent \textbf{Consistency Regularization}
These methods exploit the assumption that predictions for different perturbed versions of a sample should be consistent~\cite{li2019decoupled, Sajjadi_NIPS16}. One way to operationalise this idea is to produce several augmented versions of a given unlabeled image, then apply a loss to ensure that the predictions for all such versions are consistent. Inspired by recent methods~\cite{Sohn_fixmatch20, Berthelot_MixMatch19}, we make use of two types of augmentations, namely, weak augmentations $\mathcal{A}_w(.)$ and strong augmentations $\mathcal{A}_s(.)$, where the notion of intensity relates to how perturbing an augmentation is to an image.

\smallskip
\noindent \textbf{Pseudo-labeling}
\label{pseudo_background}
These methods rely on producing synthetic labels for unlabeled data which are then used to retrain the model. Recent alternative variations of pseudo-labeling~\cite{lee2013_pseudo, Berthelot_MixMatch19, Sohn_fixmatch20} can broadly be formalized as methods trying to account for unlabeled data by minimizing the following objective, 
\vspace{-3mm}
\begin{align}
    &\mathcal{L}(\Vec{\theta}) = \frac{-1}{\mu \cdot n}\sum_{j=1}^{\mu n}\eta_j \log p(y=\hat{y}_j\,|\,\Vec{u}_j, \Vec{\theta}) , \label{eqn:pl_formula}
\end{align}
where $\Vec{\theta}$ represents learnable model parameters, $\hat{\Vec{y}}$ denotes the pseudo-label, and $\eta$ is an arbitrary function. The choice of $\hat{\Vec{y}}$ and $\eta$ gives rise to different variations of pseudo-labeling methods ~\cite{lee2013_pseudo, Berthelot_MixMatch19, Sohn_fixmatch20, lee2013_pseudo}. We are particularly interested in confidence-based methods where
\vspace{-2mm}
\begin{align}
&\hat{y}_j =\arg\max_{y'} p(y=y'\,|\,\Vec{u}_j, \Vec{\theta}) \label{eqn:fmatch_argmax} \\ 
&~~~~~~~\text{and}~~ \eta_j=\mathds{1}(p(y=\hat{y}_j\,|\,\Vec{u}_j, \Vec{\theta})\geq \tau),  \nonumber
\end{align}

\noindent with $\mathds{1}$ denoting the Indicator function. In such methods, the unlabeled sample is only retained for pseudo-labeling if the model's maximum confidence score exceeds a predefined threshold, $\tau$, and the pseudo-label is then selected to be the class with the maximum score\footnote{For notation simplicity, we assume here that the $\arg\max$ in Eqn.~\ref{eqn:fmatch_argmax} produces a one-hot probability distribution}. This approach mitigates confirmation bias (\emph{see} Sec.~\ref{introduction}) by only retaining high confidence samples. Simultaneously, it encourages entropy minimization~\cite{Grandvalet_EntropMin05} whereby the model is encouraged to produce high confidence predictions on the unlabeled data. Recently, Sohn~\etal~\cite{Sohn_fixmatch20} combined such approach with consistency regularization to propose FixMatch, a method which achieves state-of-the-art results on several SSL image classification benchmarks. 

\begin{figure*}[h!]
    \centering
  \includegraphics[width=0.99\textwidth]{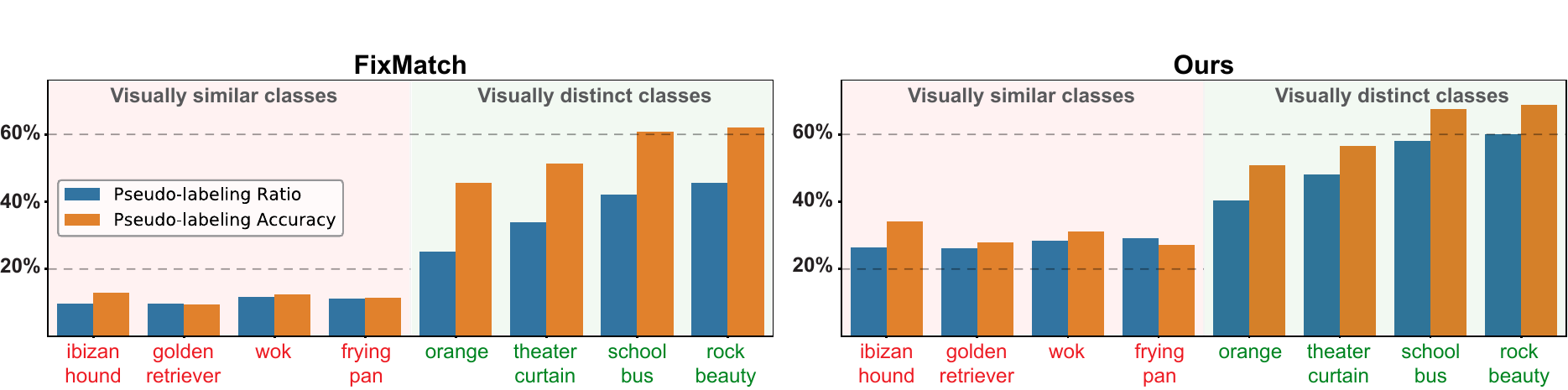}
 \caption{Confidence-based pseudo-labeling comparison between the baseline (left) and our method (right). \emph{Accuracy} values show how much, on average, pseudo-labels for a given class match the true label, while \emph{Ratio} values show the percentage of samples of a given class which are retained for pseudo-labeling (i.e. with confidence score above the threshold). The two metrics are calculated for the 4 most (red) and least (green) visually similar classes over the first 10 epochs of training.}
 \label{fig:pl_stats_comparison}
 \vspace{-5mm}
\end{figure*}

\smallskip
\noindent \textbf{Co-training}
The idea of co-training~\cite{blum1998_co_training} is to train two models with different views of the data, where each model is trained on the other’s most confident predictions. Given sufficiently diverse views of the data, this approach was shown to improve learning, as it allows the two models to learn based on their disagreements~\cite{van2020survey}. We adopt a similar strategy, albeit, we use two different views of the label rather than the data. We use the regular one-hot view as well as a distributed view (i.e. label embedding).
As we introduced in Sec.~\ref{introduction}, using a distributed view of the label grants the ability to map from the image feature space to another meaningful semantic space. This is under the assumption that the label embeddings are learnt in a way that captures semantic similarities among the labels. In~\cite{frome2013devise}, authors show how semantic information gleaned from text, in form of word embeddings~\cite{mikolov2013_word2vec}, can be exploited to enable  prediction of labels never observed during training. In this work, we combine the above ideas to propose our method.

\section{Problem Statement and Motivation}

While achieving great results, approaches that rely on pseudo-labeling share a limitation. As Eqn.~(\ref{eqn:pl_formula}) suggests, they solely rely on the model's prediction to decide about pseudo-labeling, while disregarding any prior information about possible similarities among the classes. We find in our work that visually similar classes often produce low-confidence predictions, hence are either discarded (for methods which use confidence thresholds such as FixMatch) or confused with others. This leads to class imbalance among the pseudo-labeled instances which potentially misguides SSL training. In Fig.~\ref{fig:pl_stats_comparison} - left, we demonstrate such behaviour by examining the pseudo-labeling statistics of FixMatch method. We use the true labels of the unlabeled data\footnote{Note that we have access to the true labels but they are discarded during training to emulate an SSL setting} to calculate the true accuracy of pseudo-labeling for each class. Further, we calculate the ratio of samples retained for pseudo-labeling (i.e. where the classifier confidence exceeds the threshold).  We plot these two metrics for the 4 most and least visually similar classes\footnote{We elaborate on how we identify similarity in Sec.~\ref{sec:our_method}}. We observe that visually similar concepts are chosen less frequently (i.e. less ratio) for pseudo-labeling and are often mislabeled (i.e. less accuracy) as opposed to visually distinct concepts. Motivated by this observation, we consider the label similarities as a particularly essential prior that is easy to obtain. In the subsequent sections, we will discuss how to obtain and incorporate such a prior for an improved pseudo-labeling.

\section{Our Method (SemCo)}
\label{sec:our_method}

We aim to address the issues demonstrated for visually similar classes. We build on top of recent approaches, but we additionally propose to condition pseudo-labeling on our prior knowledge of class similarities. Effectively, we enhance the model by incorporating knowledge about potential confusions based on semantic and visual similarities\footnote{We present a probabilistic interpretation of our method in the supplementary material}. 
To that end, we encode the notion of similarity among the classes using a label embeddings matrix $M \in \mathbb{R}^{K \times d}$  where each row represents a $d$-dimensional label embedding of class $k \in \{1, \cdots , K\}$\footnote{We defer the discussion on how to obtain $M$ to Sec.~\ref{extracting_emb}.}.
We further group the labels using a density-based clustering approach such as ~\cite{ester1996_DBscan} using a hyperparameter $\epsilon$ so that the number of groups are not pre-defined. Subsequently, we obtain $Q$ class groups.
 
Thereafter, we train two classifiers sharing the same backbone network (\emph{see} Fig.~\ref{fig:high_level_diagram}). The \emph{Semantic Classifier} $f_{sc}:\mathbb{R}^{\text{h} \times \text{w}} \rightarrow \mathbb{R}^d$ maps an input image closer to its corresponding label in the embedding space spanned by the rows of $M$; and the \emph{One-Hot Classifier} $f_{oh}:\mathbb{R}^{\text{h} \times \text{w}} \rightarrow \mathbb{R}^K$ maps input images to a one-hot view of the label. Note that, for brevity, we define the classifiers $f_{sc}$ and $f_{oh}$ to implicitly include the shared backbone network and its parameters.

For each of the two classifiers we minimize a supervised loss on the labeled data and an unsupervised consistency loss on the unlabeled data. Additionally, we add a loss term for co-training to allow the two classifiers to co-operate on pseudo-labeling. 

\subsection{Semantic Classifier}
\label{embedding_classifier}

For the supervised loss, we minimize the \emph{cosine loss} between the true label embedding and the predicted label embedding, 
\vspace{-3mm}
\begin{align}
\mathcal{L}^{sc}_s &= \frac{1}{n} \sum_{i=1}^{n} C(M^T \Vec{y}_i, f_{sc}(\Vec{x}_i)), \label{eqn:loss_sc_s}\\
\hfill\break
\text{with}~~C(&\Vec{z}, \Vec{z'}) = 1 - \text{CosineSim}(\Vec{z},  \Vec{z'}). \nonumber
\end{align}
\vspace{-5mm}

\begin{figure*}[h!]
 \centering
  \includegraphics[width=0.95\textwidth]{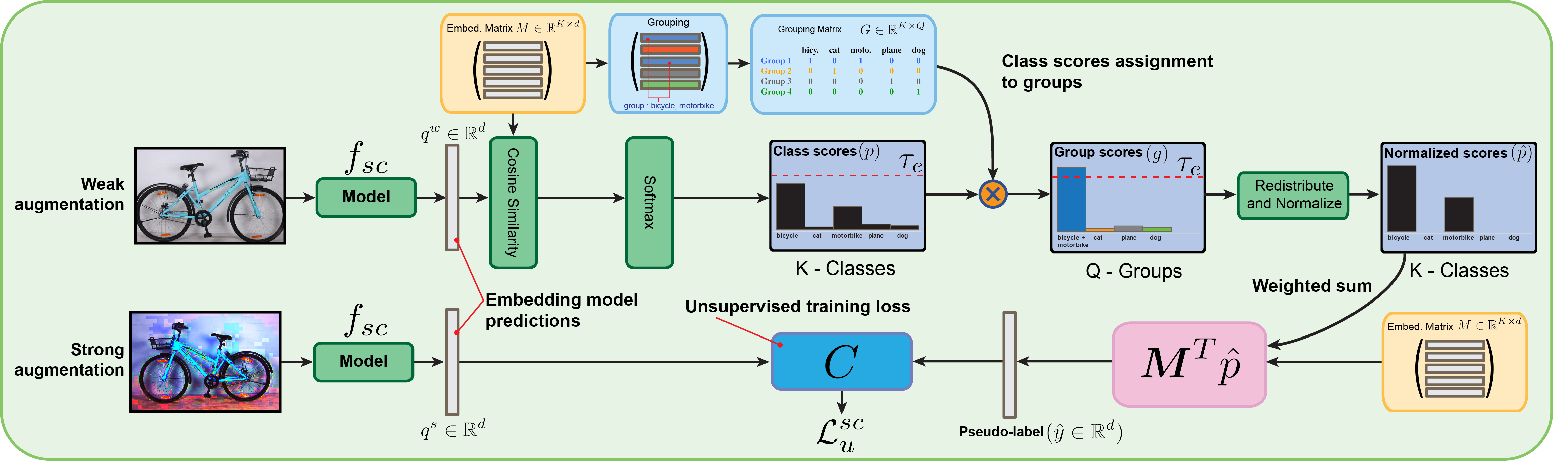}
 \caption{Unsupervised loss for the \emph{Semantic Classifier} - A weakly augmented image is used (upper path) to obtain a predicted embedding, which is then used to obtain class scores. The class scores are summed for each label group (as identified by our grouping method) to obtain group scores. If one of the group scores exceeds the threshold, it is retained for pseduo-labeling. The pseudo-label is then calculated as an average of the group members embeddings weighted by their class scores. The loss is then enforced against the predicted embedding for a strongly augmented image (lower path).}
 \vspace{-5mm}
\label{fig:mainDiag}
\end{figure*}

For the unsupervised loss (\emph{see} Fig.~\ref{fig:mainDiag}), we draw inspiration from FixMatch~\cite{Sohn_fixmatch20}, where we use a weakly augmented version of the image to obtain a pseudo-label and enforce that against the model's prediction for a strongly augmented version of the same image. Specifically, for an unlabeled image $\Vec{u}_j$, we obtain the predicted embedding for a weakly augmented version of the image: $\Vec{q}_j = f_{sc}(\mathcal{A}_w(\Vec{u_j}))$. Then we calculate class scores, $\Vec{p}_j = p(\Vec{y}_j | \Vec{u}_j)$ by normalizing the vector-wise cosine similarity between $\Vec{q}_j$ and $M$. 


\smallskip
\noindent Unlike FixMatch, we consider an unlabeled sample for pseudo labeling if the prediction score for one of the class groups (as opposed to the individual classes) exceeds a predefined threshold ($\tau_e$). To elaborate, referring to Fig.~\ref{fig:mainDiag}, due to the visual similarity between ``\emph{bicycle}'' and ``\emph{motorbike}'', the class scores for each of them, individually, is falling below the threshold. However, since they are both identified as ``visually similar'' based on clustering their embeddings, their scores are added first before applying the threshold. The combined score exceeds the threshold so the sample is retained for pseudo-labeling where the pseudo-label is calculated as the average of the ``\emph{bicycle}'' and ``\emph{motorbike}'' embeddings weighted by their normalized class prediction scores.

\noindent To put it formally, to obtain the score for a given class label group, we sum the normalized class scores of all its members (where the membership is defined based on clustering $M$). This gives rise to our group scores $\Vec{g}_j$\footnote{$\Vec{g}_j$ is calculated as the inner product of $\Vec{p}_j$ with the cluster assignment matrix}.

\noindent Thereafter, we apply our mask to select samples for pseudo-labeling as per,
\begin{align}
\eta^{sc}_j = \mathds{1}(\max(\Vec{g}_j) \geq \tau_e) \label{eqn:mask_sc}.
\end{align}
If a sample is selected for pseudo-labeling, we obtain a pseudo-label embedding ($\hat{\Vec{y}}_j$)\footnote{$\hat{\Vec{y}}_j$ is calculated as the inner product of $M$ with the normalised class scores.} for such sample as a weighted average of the group members embedding, where we weigh the average based on the original class scores $\Vec{p}_j$. 

\noindent Consequently, we apply the loss against the embedding prediction of a strongly augmented version of $\Vec{u}_j$ as per,
\vspace{-3mm}
\begin{align}
\mathcal{L}^{sc}_u& =\frac{1}{\mu \cdot n} \sum_{j=1}^{\mu \cdot n} C(\hat{\Vec{y}}_j, f_{sc}(\mathcal{A}_s(\Vec{u}_j))) \cdot \eta^{sc}_j. \label{eqn:loss_sc_u}
\end{align}

\subsection{One-Hot Classifier}
For the \emph{One-Hot Classifier}, we follow the same procedure as the \emph{Semantic Classifier} with two crucial differences:  1) we use \emph{cross-entropy loss} instead of \emph{cosine loss}, and 2) we don't apply label grouping before comparing with the confidence threshold.  We note here that this classifier operates in a similar way to FixMatch, yet we include the loss equations for completeness. By analogy, the supervised loss is calculated as,
\begin{align}
\mathcal{L}^{oh}_s = \frac{1}{n} \sum_{i=1}^{n} H(\Vec{y}_i, f_{oh}((\Vec{x}_i))). \label{eqn:loss_oh_s}
\end{align}

\noindent Here, we use $H$ to represent the cross-entropy loss function. To this end, the unsupervised loss is formulated as,

\vspace{-5mm}
\begin{align}
\mathcal{L}^{oh}_u = \frac{1}{\mu \cdot n} \sum_{j=1}^{\mu \cdot n} H(\hat{\Vec{y}}_j, f_{oh}(\mathcal{A}_s(\Vec{u}_j))) \cdot \eta^{oh}_j \label{loss_oh_u},
\end{align}
where $\hat{\Vec{y}}_j = \argmax \mathcal{A}_w(\Vec{u_j})$; $\eta^{oh}_j = \mathds{1}(\max(\hat{\Vec{y}}_j) \geq \tau_o)$.
We note that our motivation behind using two different types of loss functions is related to the concept of co-training. An assumption underlying the success of co-training is to ensure that the learners are sufficiently diverse so that they learn better based on their different views~\cite{van2020survey}. We validate such choice in our ablations (Sec.~\ref{ablation}) 
\subsection{Co-training Loss}
This loss is meant to enable both classifiers to learn from each other. The intuition is that due to each classifier's different view of the labels, they will each be confident about different samples of the unlabeled data. We exploit that by retaining a sample for pseudo-labeling if either of the classifiers is confident about its prediction. In case the two classifiers disagree about a sample (i.e. they are both confident about two different labels), the sample is included twice in the loss, once with each of the two pseudo-labels. We experimented with another approach, where in such event, the sample gets discarded but it degraded the performance. We conjecture that it's because the former approach encourages the two classifiers to be consistent while the latter completely ignores the confusion event. Formally, we define the co-training loss as,
\vspace{-5mm}

\begin{align}
\mathcal{L}_{co}& =\frac{1}{\mu \cdot n} \sum_{j=1}^{\mu \cdot n} C(M^T \hat{\Vec{y}}_j, f_{sc}(\mathcal{A}_s(\Vec{u}_j))) \cdot \eta^{oh}_j \nonumber\\
&\qquad\qquad + H(\arg\max(\Vec{p}_j), f_{oh}(\mathcal{A}_s(\Vec{u}_j))) \cdot \eta^{sc}_j \label{eqn:loss_co}
\end{align}

\subsection{Total Loss}
We now define our final training loss function by combining all five losses (Eqns.~\ref{eqn:loss_sc_s}, and \ref{eqn:loss_sc_u} to \ref{eqn:loss_co}) as per,
\vspace{-2mm}

\begin{align}
\vspace{-2mm}
\mathcal{L}_{total} = \mathcal{L}^{sc}_s + \mathcal{L}^{oh}_s + \lambda_u (\mathcal{L}^{sc}_u + \mathcal{L}^{oh}_u) + \lambda_{co} \mathcal{L}_{co}.
\end{align}

\noindent Here, $\lambda_u$ and $\lambda_{co}$ are fixed scalar weights to modulate the contribution of the unsupervised loss and co-training loss, respectively. 

\subsection{Extracting Label Semantics}
\label{extracting_emb}
In this section, we propose two alternatives to obtain the label embedding matrix $M$ which establishes our prior on the visual similarity among the classes. 

\smallskip
\noindent \textbf{Using Knowledge Graphs}
In cases where the class labels are semantically meaningful, we make use of the ConceptNet knowledge graph~\cite{speer2016_conceptnet} together with GloVe~\cite{pennington2014_glove} and word2vec~\cite{mikolov2013_word2vec} distributional embeddings as the basis for obtaining the distributed label embeddings. ConceptNet is a multilingual knowledge graph that connects words of natural language with labeled, weighted relations. Since our main goal is to obtain label embeddings which capture visual similarity, we filter the graph to only retain relations which imply such similarity. Specifically, we retain any nodes which share the following relations: \emph{SimilarTo}, \emph{InstanceOf}, \emph{IsA}, \emph{FormOf}, \emph{Synonym}, \emph{EtymologicallyRelatedTo}, \emph{DefinedAs}. A detailed description of such relations and examples thereof can be found in the ConceptNet documentation\footnote{\url{https://github.com/commonsense/
conceptnet5/wiki/Relations}}. On the other hand, GloVe and word2vec are two prominent sets of word embeddings, the former is trained on 840 billion words of the Common Crawl~\cite{pennington2014_glove}, while the latter is trained on 100 billion words of Google News~\cite{mikolov2013_word2vec}. The two sets capture the distributional similarity among the different words but don’t necessarily capture visual similarity. For example, ``\emph{cat}'' and ``\emph{dog}'' usually appear in similar contexts (being both animal pets) so they would have a relatively similar GloVe (or word2vec) word embedding even though they are not visually similar. Combining the distributional embeddings with the ConceptNet filtered graph allows us to address this problem: we follow a procedure similar to the authors in~\cite{speer2016_conceptnet} to retrofit the distributional embeddings with the filtered knowledge graph. Retrofitting~\cite{faruqui2014_retrofitting} is a process which adjusts a word embedding matrix based on a knowledge graph by optimizing an objective function which tries to find for each term a new vector close to the vector’s original value but also close to the term neighbours in the graph. Since the retained relations in the graph are those which implies visual similarity, this retrofitting results in a new hybrid set of embeddings which captures distributional similarity but also correlates well with visual similarity. Finally, to handle class labels which are not present in the embeddings vocabulary, we implement a fall-out strategy to find the most reasonable alternative. We provide further description of the retrofitting process and we show a qualitative comparison to demonstrate its effectiveness in the supplementary material. We also provide further details of the fall-out strategy.

\smallskip
\noindent \textbf{Using Class Attributes Annotations}
In cases where the class labels are not semantically meaningful, a viable alternative is to use manually annotated class attributes. We demonstrate (Sec.~\ref{experiments}) that by using attributes annotation of CUB-200~\cite{cub200} fine-grained dataset as our $M$ matrix, we achieve significant gains against the baseline. Considering that the cost of annotating attributes is expected to be cheaper than annotating data instances, we propose class attributes annotation as a possible alternative.

%% file: 3Experiments.tex
\section{Experiments}
\label{experiments}

\begin{table*}[h!]
\Large
\caption{Error rates for CIFAR-10, CIFAR-100 and Mini-ImageNet. We report results for two different values of $\mu$ - i.e. ratio between unlabeled and labeled data in a mini-batch, for our method and FixMatch. $\dagger$ denotes that the results reported are using the same codebase. $*$ denotes that the result is based on using CNN-13 model. We report the mean and standard deviation across 3 different splits of labeled data for each experiment.}
\centering
\label{main_results_table}
\resizebox{\textwidth}{!}{%
\begin{tabular}{llcclccclccc}
\hline
                       &  & \multicolumn{2}{c}{\textbf{CIFAR-10}}    &  & \multicolumn{3}{c}{\textbf{CIFAR-100}}                          &  & \multicolumn{3}{c}{\textbf{Mini-ImageNet}}                                  \\ \cline{1-1} \cline{3-4} \cline{6-8} \cline{10-12} 
Total Labelled Samples &  & 250                  & 4000              &  & 2500                & 4000                & 10000               &  & 1000                & 4000                & 10000                           \\ \cline{1-1} \cline{3-4} \cline{6-8} \cline{10-12} 
Pseudo-labeling~\cite{lee2013_pseudo}        &  & 49.78$\pm$0.43           & 16.09$\pm$0.28        &  & -                   & -                   & -                   &  & -                   & -                   & -                               \\
Mean teacher~\cite{mean_teacher}           &  & 32.32$\pm$2.30           & 9.19$\pm$0.19         &  & -                   & -                   & -                   &  & -                   & 72.51$\pm$0.22          & 57.55$\pm$1.11 \\
UDA~\cite{xie2019_uda}                    &  & 8.82$\pm$1.08            & 4.88$\pm$0.18         &  & 33.13$\pm$0.22          & -                   & 24.50$\pm$0.25          &  & -                   & -                   & -                               \\
Label Propagation~\cite{label_prop}      &  & -                    & 12.69$\pm$0.29\textsuperscript{$*$}     &  & -                   & -                   & -                   &  & -                   & 70.29$\pm$0.81          & 57.58$\pm$1.47 \\
PLCB~\cite{arazo_pseudo}                   &  & 24.81$\pm$5.35         & 6.28$\pm$0.30        &  & -                   & 37.55$\pm$1.09\textsuperscript{$*$}       & 32.15$\pm$0.50\textsuperscript{$*$}       &  & -                   & 56.49$\pm$0.51          & 46.08$\pm$0.11 \\
MixMatch\textsuperscript{$\dagger$}~\cite{Berthelot_MixMatch19}            &  & 11.29$\pm$0.75           & 6.24$\pm$0.07         &  & 39.70$\pm$0.27          & -                   & 28.59$\pm$0.31          &  & 60.97$\pm$0.31                   & 49.79$\pm$0.11                   & 44.27$\pm$0.23                               \\
FixMatch\textsuperscript{$\dagger$}($\mu=3$)~\cite{Sohn_fixmatch20}    &  & 5.78$\pm$0.23 & 4.52$\pm$0.01      &  & 38.45$\pm$0.51          & 32.22$\pm$0.21          & 28.42$\pm$0.09          &  & 66.23$\pm$1.13          & 59.73$\pm$5.45          & 44.66$\pm$0.12                      \\
FixMatch\textsuperscript{$\dagger$} ($\mu=7$)   &  & \textbf{4.55$\pm$0.12}   & 4.49$\pm$0.05         &  & 33.64$\pm$0.07          & 31.27$\pm$1.30           & 26.13$\pm$0.18          &  & 60.97$\pm$0.31          & 49.79$\pm$0.11          & 44.27$\pm$0.23                      \\ \hline
Ours (SemCo)\textsuperscript{$\dagger$} ($\mu=3$)       &  & 5.87$\pm$0.31           & 4.43$\pm$0.01        &  & 33.80$\pm$0.57           & 29.40$\pm$0.18           & 25.07$\pm$0.04          &  & \textbf{55.35$\pm$0.71} & \textbf{46.01$\pm$0.93} & \textbf{41.25$\pm$0.76}            \\
Ours (SemCo)\textsuperscript{$\dagger$} ($\mu=7$)       &  & 5.12$\pm$0.27            & \textbf{3.80$\pm$0.08} &  & \textbf{31.93$\pm$0.01} & \textbf{28.61$\pm$0.23} & \textbf{24.45$\pm$0.12} &  & 59.35$\pm$0.23          & 49.46$\pm$2.20           & 42.78$\pm$0.35                      \\ \hline
\end{tabular}%
}
\end{table*}

To evaluate SemCo, we compare it to various recent SSL baselines on 3 standard benchmarks (CIFAR-10~\cite{cifar100}, CIFAR-100~\cite{cifar100}, Mini-ImageNet~\cite{mini_imagenet}).  Further, we experiment on 2 other datasets: CUB-200~\cite{cub200}, to test SemCo on fine-grained tasks; and DomainNet~\cite{peng2019moment_domainnet}, to verify its performance on larger more complex images.

\subsection{Datasets}

\smallskip
\noindent \textbf{CIFAR-10/100}
Both datasets comprise natural images of 10, and 100 classes respectively. Their training set consists of 50k images while the test set consists of 10k images. All the images have a fixed resolution of 32x32. We conduct three different experiments on each of them with varying amounts of labeled data as shown in Table \ref{main_results_table}.

\smallskip
\noindent \textbf{Mini-ImageNet}
A subset of the well-known ImageNet~\cite{russakovsky2015imagenet_large}. It consists of 100 classes with 600 images per class (84x84 each). We use the same train/test split used by \cite{label_prop} and we create splits for 40 and 100 labeled images per class to enable comparing with the baseline systems. However, we also experiment with 10 images per class to test SemCo in the low data regime. 

\smallskip
\noindent \textbf{CUB-200}
A fine-grained  image classification dataset comprising 11k images from 200 different types of birds annotated with 312 attributes per class. We experiment with 5 and 10 images per class corresponding to 1000 and 2000 total labeled data.

\smallskip
\noindent \textbf{DomainNet}
The dataset contains 345 classes of images coming from six domains: \emph{Clipart}, \emph{Infograph}, \emph{Painting}, \emph{Quickdraw}, \emph{Real}, and \emph{Sketch}. We report results only on the \emph{Real} domain to evaluate how our method works on larger more complex datasets.

\subsection{Experimental Setup}
Across all experiments, we  follow the standard approach where we randomly select a certain number of samples to represent our labeled set and ignore the labels of the remaining samples and use them to form our unlabeled set. For the standard benchmarks, we compare our results to various existing baselines~\cite{lee2013_pseudo, mean_teacher, Berthelot_MixMatch19, xie2019_uda, label_prop, arazo_pseudo, Sohn_fixmatch20}, which employ consistency regularization and/or pseudo-labeling (see Sec.~\ref{related_work}).
For the two other datasets, we only compare with FixMatch, being the most similar to our solution and the closest in performance. 

Since SemCo bears the most resemblance with \cite{Sohn_fixmatch20}, \cite{Berthelot_MixMatch19}, and \cite{Berthelot_RemixMatch19}, we follow the recommendation of Oliver~\etal~\cite{oliver_realistic} for a realistic comparison: we integrate the implementation of their methods\footnote{we don't report results for~\cite{Berthelot_RemixMatch19} due to adaptation difficulties} into our codebase and use the unified codebase to conduct all the experiments. As for the other baselines, we report the results as mentioned in the original papers, provided that the result is based on the same model architecture we use.  We use WideResnet-28-2~\cite{zagoruyko2016_wideresnet} for CIFAR-10/100, Resnet-18~\cite{he2016_resnet} for Mini-ImageNet, and Resnet50 for CUB-200 and DomainNet. Additionally, we attach a fully connected layer to the encoder output to act as our \emph{Semantic Classifier} (\emph{see} Fig.~\ref{fig:high_level_diagram}). We train our model end-to-end along with the backbone network.

Unless otherwise specified, we use the same hyperparameters for all our experiments. These were tuned on a validation set for a single experiment (CIFAR100 - 2500 labels) and then fixed across all other experiments. In general, we found that our model is not sensitive to the values of $\lambda_u$ and $\lambda_{co}$. Values between 0.5 - 1 all yielded similar performance, albeit, smaller values of $\lambda_u$ slightly slowed convergence. Further, we found that our model is mostly sensitive to $\epsilon$ - the label grouping parameter and hence it was the only parameter we tuned separately for each dataset (\emph{see} supplements for the full list of hyperparameters).

Since the labels of all datasets are semantically meaningful, we use their retrofitted embeddings (see Sec.~\ref{extracting_emb}) as targets for our \emph{Semantic Classifier}. The only exception is for CUB-200 where we use human annotated attributes as targets to test our alternative proposal in Sec.~\ref{extracting_emb}. We start from the 312 dimensional class attributes given in~\cite{cub200} and reduce their dimensionality to 128 using PCA~\cite{wold1987principal_pca}. We, then, use the obtained class attributes matrix as targets for our \emph{Semantic Classifier}. 

\subsection{Main Results}
We report standard benchmarks results in Table \ref{main_results_table} and CUB-200 and DomainNet results in Table~\ref{cub200_table}. We observe that SemCo outperforms all the baselines with a large margin across the different datasets and amounts of labeled data (except for one case). Notably, SemCo achieves an average error rate of 55.35\% on Mini-ImageNet with 1000 labels (i.e. 10 samples per class). This is almost 5.6\% improvement compared to the closest baseline. We note here that Mini-ImageNet classes include 13 different species of dogs which share many visual similarities. SemCo grouped 7 of these classes into a single group based on clustering their label embeddings.
To understand why the performance degrades on CIFAR10 (250 labeled samples), we looked into the clustering results for CIFAR-10 class embeddings. We observed that the 10 classes were deemed visually distinct by our clustering component, leading to one-member clusters for all 10 classes. The two above results align with our original hypothesis that SemCo is particularly useful when there are visually similar concepts among the classes. Evidently, in such cases, using label grouping in conjunction with our co-training routine helps improving the pseudo-labeling quality. This is also consistent with the pseudo-labeling statistics shown in Fig.~\ref{fig:pl_stats_comparison} - right, where we can see that SemCo significantly improves both the quality and the quantity of pseudo-labeled data. 

\smallskip
\noindent \textbf{Ratio of unlabeled data}
We observe that SemCo achieves better results with less batch size as opposed to the baseline. As shown in Table \ref{main_results_table}, we experiment with different values of $\mu$, which defines the ratio between unlabeled and labeled data in each training batch. We find that our method consistently achieves better results even when using less unlabeled data. For example, for CIFAR-100 (4000 labels), we achieve less average error rate (29.4\%) with $\mu=3$ than FixMatch does with $\mu=7$ (31.27\%). More notably, we achieve 13\% improvement on Mini-ImageNet (4000 labels) when fixing $\mu$ to $3$.

\begin{figure*}[h!]
 \centering
  \includegraphics[width=0.97\textwidth]{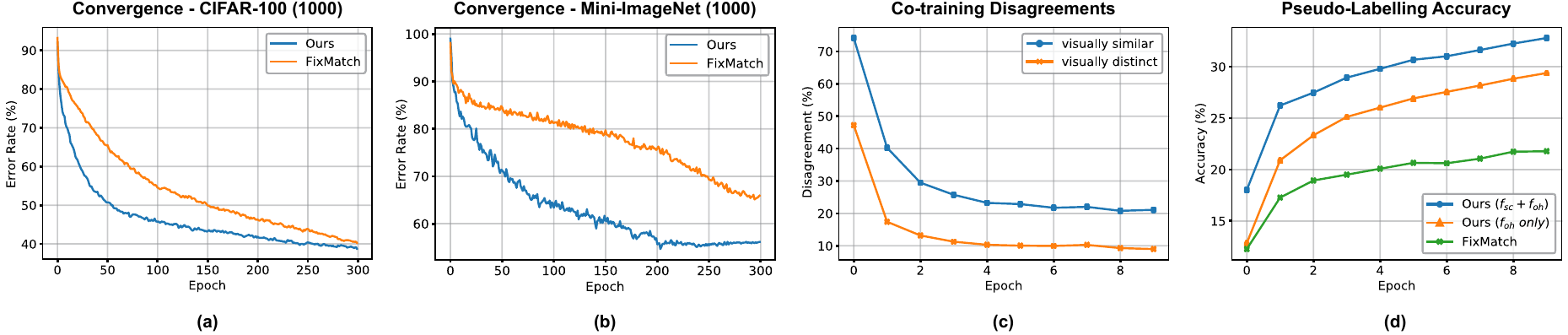}
 \caption{Experimental analysis plots showing: (a,b): Convergence trends of our method and the baseline for CIFAR-100 (a) and Mini-ImageNet (b) with 1000 labeled examples. (c,d): Co-training analysis plots showing the disagreements between our two classifiers for visually similar and distinct classes (c) and the associated pseudo-labeling accuracies (d). The co-training plots are spanning only the first 10 epochs of training.}
 \label{fig:analysis}
 \vspace{-5mm}
\end{figure*}

\smallskip
\noindent \textbf{Co-training Analysis}
Further, we investigate the effectiveness of our co-training routine. We use the same experimental setup of capturing pseudo-label metrics (\emph{see} Fig.~\ref{fig:pl_stats_comparison}), but this time, we monitor the rate of disagreement on pseudo-labels among the two classifiers $f_{sc}$ and $f_{oh}$ (i.e. percentage of time the two classifiers are confident about different pseudo-labels for the same unlabeled sample). In Fig.~\ref{fig:analysis}~c, we report disagreement curves for the same 8 classes shown in Fig.~\ref{fig:pl_stats_comparison}. As the training progresses, we track the pseudo-labeling accuracy (Fig.~\ref{fig:analysis}~d) for each of: 1) our classifier ensemble, 2) our one-hot classifier $f_{oh}$, and 3) the baseline (FixMatch). Note that $f_{oh}$ is using the same method for pseudo-labeling (i.e. confidence threshold on non-grouped labels) as FixMatch. We find that at the beginning of the training, both classifiers highly disagree about pseudo-labeling, especially for visually similar classes. As the training progresses, we witness a sharp reduction in disagreements coupled with an increase in accuracy for both classifiers. Interestingly, we find that the accuracy of our $f_{oh}$ is considerably higher than FixMatch although both are using the same basis for pseudo-labeling. This demonstrates the success of co-training in making pseudo-labeling consistent and accurate by leveraging the co-operation between the two classifiers. 

\smallskip
\noindent \textbf{Convergence Speed}
In Fig.~\ref{fig:analysis}~a~and~b, we study the convergence plots on Mini-ImageNet and CIFAR-100 (1000 labels). We observe that SemCo achieves the best performance of the baseline with significantly less training iterations. This can be explained through Fig.~\ref{fig:analysis}~d: the higher accuracy of pseudo-labeling in the early phase of the training helps better guide the learning and thereby translates to faster convergence.

\begin{table}[h!]
\caption{Error rates on CUB-200 dataset and DomainNet Real. Errors are reported based on 1 split for each of the amounts of labeled data. Poor baseline results are omitted.}
\label{cub200_table}
\small
\centering
\resizebox{0.4\textwidth}{!}{%
\begin{tabular}{llrr}
\hline
\textbf{CUB-200}                    &  & \multicolumn{2}{l}{Total Labeled Samples} \\ \cline{3-4} 
Method              &  & 1000                    & 2000            \\ \cline{1-1} \cline{3-4} 
Supervised baseline &  & \multicolumn{1}{c}{-}   & 70.11           \\ \hline
FixMatch            &  & 84.35                   & 72.15           \\
Ours (SemCo)                &  & \textbf{79.44}          & \textbf{66.76}  \\ \hline
\end{tabular}%
}
\end{table}

\begin{table}[h!]
\small
\centering
\resizebox{0.4\textwidth}{!}{%
\begin{tabular}{lccc}
\hline
\textbf{DomainNet Real}                    & \multicolumn{1}{l}{} & \multicolumn{2}{l}{Total Labeled Samples} \\ \cline{3-4} 
Method              &                      & 6900                & 10350               \\ \cline{1-1} \cline{3-4} 
Supervised baseline &                      & 47.9                & 45.2                \\ \hline
FixMatch            &                      & 41.34               & 39.04               \\
Ours (SemCo)               &                      & \textbf{35.32}      & \textbf{32.89}      \\ \hline
\end{tabular}%
}
\end{table}

\subsection{Ablation}
\label{ablation}
We are interested in isolating the contribution of each of the three key components of SemCo towards the witnessed performance gain.

\smallskip
\noindent \textbf{Label Grouping \& Co-training}
In Table \ref{ablation_table}, we investigate the effect of label grouping (by controlling our clustering hyperparameter $\epsilon$), and co-training (by toggling $\lambda_{co}$). We observe that both components are almost equally important towards the witnessed performance gain. However, co-training seems to provide a slight advantage over label grouping in both experiments.

\begin{table}[h!]
\large
\centering
\caption{Error Rates for different settings of Co-training and Label Grouping}
\label{ablation_table}
\resizebox{0.45\textwidth}{!}{%
\begin{tabular}{cclcc}
\hline
\multicolumn{1}{l}{} & \multicolumn{1}{l}{} &  & \textbf{\begin{tabular}[c]{@{}c@{}}Mini-ImageNet\\ 1000\end{tabular}} & \textbf{\begin{tabular}[c]{@{}c@{}}CIFAR-100\\ 2500\end{tabular}} \\ \cline{1-2} \cline{4-5} 
Label Grouping       & Co-training          &  & \multicolumn{2}{c}{Error Rate}                                                                                                            \\ \cline{1-2} \cline{4-5} 
$\checkmark$                   & $\checkmark$                   &  & 55.35                                                                & 31.93                                                             \\
-                    & $\checkmark$                   &  & 59.60                                                                & 33.09                                                             \\
$\checkmark$                   & -                    &  & 60.39                                                                 & 33.19                                                             \\
-                    & -                    &  & 62.16                                                                 & 34.25                                                             \\ \hline
\end{tabular}%

}
\end{table}

\smallskip
\noindent \textbf{Label Embeddings as Training Targets}
We experiment on CIFAR-100 and Mini-ImageNet in another setting where we use the one-hot target for both our classifiers. In such case, the only difference between the two classifiers is that the \emph{Semantic Classifier} implements label grouping while the \emph{One-Hot Classifier} does not. In Table~\ref{ablation_table_emb}, we observe a significant decrease in performance when using the one-hot view for both the classifiers. This strongly supports our hypothesis that co-training with different views of the label does indeed help the learning.

\begin{table}[h!]
\large
\caption{Error Rates when using Embedding Targets versus One-Hot Targets for our \emph{Semantic Classifier}, reported on CIFAR-100 and Mini-ImageNet}
\label{ablation_table_emb}
\centering
\resizebox{0.4\textwidth}{!}{%
\begin{tabular}{lc c}
\cline{1-3}
                     & Embeddings Target & One-Hot Target                \\ \hline
CIFAR-100 (2500)     & 31.93             & 33.33 \\
Mini-ImageNet (1000) & 55.35             & 60.33 \\ \hline
\end{tabular}%
}
\end{table}

%% file: 4RelatedWork.tex
\section{Related Work}
\label{related_work}
Since the seminal ``$\Pi$-model'' \cite{rasmus2015semi}, consistency regularization and pseudo-labelling SSL solutions have seen improvements in the consistency propagation~\cite{mean_teacher,Laine_temporal16}, 
and approaches for generating diverse views~\cite{french2020milking,kuo2020featmatch,miyato2018virtual}.
For instance, the Mean-Teacher~\cite{mean_teacher} proposes a teacher model, where its parameters are updated according to an exponential moving average (EMA) rule. Temporal Ensembling~\cite{Laine_temporal16}, maintains an EMA over the predictions for the consistency loss computation. French~\etal~\cite{french2020milking} explore a masking based approach for generating diverse views. The interpolation training given in \cite{verma2019interpolation} computes the consistency between interpolated views of unlabelled instances. 
Miyato~\etal~\cite{miyato2018virtual} explores using adversarial methods for perturbation to create diverse views.

\vspace{-2.5pt}
Our solution bears a lot of similarity to FixMatch and ReMixMatch~\cite{Sohn_fixmatch20,Berthelot_RemixMatch19} where the main idea is to use a weakly augmented image to obtain a pseudo-label then enforce it against the model's prediciton for a strongly augmented one. ReMixMatch uses a soft pseudo-label via sharpening, while FixMatch uses a hard label based on confidence. Our method compliments theirs by also conditioning on label semantics while pseudo-labeling.

Using label semantics to benefit learning is not a new idea, prior knowledge from language models~\cite{pennington2014_glove}, graph embeddings~\cite{speer2016_conceptnet}, and attribute vector representaions~\cite{cub200} has helped pushing the performance of computer vision models. In the pioneering work, DeViSE~\cite{frome2013devise} showed distributed label representations derived from unannotated text are helpful for image classification. They also extend their solution to Zero-Shot Learning (ZSL)~\cite{xian2017zero}. Ye~\etal~\cite{ye2018learning_embedding} proposes distributed labels for Few-shot learning. Such label representations are informative to even generate descriptive representations for classification~\cite{xian2018feature} and has become the backbone representation for ZSL~\cite{kodirov2017semantic,Liu_2020_CVPR}. To this end, literature provides explorations on learning visual-semantic embedding spaces with better discriminative properties~\cite{wu2019unified,hubert2017learning}.
However, to our best knowledge the capacity of such label representations has not been explored for SSL.
\vspace{-5mm}

%% file: 5Conclusion.tex
\section{Conclusion}
In this paper, we have introduced a novel semi-supervised learning approach leveraging class label semantics and co-training for more effective and efficient  learning. We  operationalize this approach for  image classification, and demonstrate that it leads to significant gains. We believe the key ingredients of our approach are general and can be extended to  supervised and unsupervised learning settings, which we will explore in the future work. 

\section*{Acknowledgement}
\noindent This work was partly supported by DARPA’s Learning with Less Labeling (LwLL) program under agreement FA8750-19-2-0501 and by the Australian Government Research Training Program (RTP) Scholarship.

%% file: 6Supplementary.tex
\setcounter{section}{0}
\setcounter{equation}{0}
\renewcommand{\thesection}{\Alph{section}}
\newpage
\newpage
\section{Probabilistic Interpretation of SemCo}
In this section, we provide a probabilistic interpretation of our method described in Sec.~\ref{sec:our_method}.\footnote{All section, table, and figure references are following the original paper numbering.}

\smallskip
\noindent We start by recalling the general form of recent pseudo-labeling methods captured by the below formalization for the unsupervised loss:
\begin{align}
    &\mathcal{L}(\Vec{\theta}) = \frac{-1}{\mu \cdot n}\sum_{j=1}^{\mu n}\eta_j \log p(y=\hat{y}_j\,|\,\Vec{u}_j, \Vec{\theta}) , \label{eqn:pl_formula_supplement}
\end{align}
which is typically added to the loss for the labeled data.
\noindent To reiterate, we use $\Vec{\theta}$ to represent learnable model parameters. For an unlabelled input $\Vec{u}_i$, $\hat{\Vec{y}}$ denotes the pseudo-label and $\eta$ is an arbitrary function. The choice of $\hat{y}$ and $\eta$ gives rise to three distinct variations of pseudo-labeling that can be written as, 
\begin{align}
&\hat{y}_j =\arg\max_{y'} p(y=y'\,|\,\Vec{u}_j, \Vec{\theta}) ~\text{with}~ \eta_j=1, \label{eqn1:naive_pl}\\
&\hat{y}_j =\frac{\exp(f_{\ell}(\Vec{u}_j)/T)}{\sum_k\exp(f_k(\Vec{u}_j)/T)} ~\text{with}~ \eta_j=1, \label{eqn2:sharpening_pl} ~\text{and}\\
&\hat{y}_j =\arg\max_{y'} p(y=y'\,|\,\Vec{u}_j, \Vec{\theta}) \label{eqn3:fmatch_pl} \\\nonumber
&~~~~~~~\text{with}~~ \eta_j=\mathds{1}(p(y=\hat{y}_j\,|\,\Vec{u}_j, \Vec{\theta})\geq \tau).  \nonumber
\end{align}
\noindent The first approach (Eqn.~\ref{eqn1:naive_pl}) corresponds to naive pseudo-labeling where the class with the highest confidence is used as a pseudo-label regardless of its score, while the second (Eqn.~\ref{eqn2:sharpening_pl}) improves on that by employing temperature sharpening ~\cite{Berthelot_MixMatch19} via $T$. For brevity, we use $f_{\ell}$ is the onehot logit for class $\ell$. Sharpening the pseudo-label implicitly encourages entropy minimizaton~\cite{Grandvalet_EntropMin05} whereby the classifier is encouraged to produce high confidence predictions on the unlabeled data. The third approach (Eqn.~\ref{eqn3:fmatch_pl}) adopted in ~\cite{Sohn_fixmatch20, lee2013_pseudo} where the unlabeled sample is only retained for pseudo-labeling if the max confidence score exceeds a predefined threshold, $\tau$. This simultaneously encourages entropy minimization\footnote{Note that it is equivalent to sharpening with $T \rightarrow 0$} and decrease confirmation bias (see Sec.~\ref{introduction}) by only retaining high confidence samples.

As opposed to above methods, we additionally propose to take a multi-view approach in which we have $y'$ as different representation of the label as well as a grouping of the similar labels potentially (but not necessarily) obtained from it. We consider this additional label to be conditionally independent of the one-hot representation of the label. 
Specifically, instead of Eqn.~\ref{eqn:pl_formula_supplement}, we propose to minimize the objective,
{\small
\begin{align}
    \mathcal{L}(\Vec{\theta}) &= \frac{-1}{\mu \cdot n}\sum_j \log\left(\, p(y=\hat{y}_j\,|\,\Vec{u}_j, \Vec{\theta})p(y'=\hat{y'}_j\,|\,\Vec{u}_j, \Vec{\theta})^{\eta_j} \,\right),\nonumber \\
    & p(y'=\hat{y'}_j\,|\,\Vec{u}_j, \Vec{\theta}) =\sum_c 
    \underbrace{p(y'=\hat{y'}_j\,|\,c, \Vec{u}_j, \Vec{\theta})}_{\text{additional classification}}\underbrace{p(c\,|\,\mathcal{Y}', \Vec{\theta})}_{\text{grouped semantics}},
    \label{eq:loss1}
\end{align}}
where $\mathcal{Y}'$ denotes the collection of the labels which we consider to be conditionally independent of the conventional one-hot label. We use the density-based clustering to calculate $p(c\,|\,\mathcal{Y}', \Vec{\theta})$.
Then by using Jensen's inequality, we have the following as the upper-bound on the loss in Eqn.~\ref{eq:loss1}:
\begin{align}
    \mathcal{L}(\Vec{\theta}) &\geq \frac{-1}{\mu \cdot n}\sum_j \eta_j\Big[\log\left(\, p(y=\hat{y}_j\,|\,\Vec{u}_j, \Vec{\theta})\,\right)\nonumber \\
    & \qquad+\sum_c \underbrace{\log\left(\,p(y'=\hat{y'}_j\,|\,c, \Vec{u}_j, \Vec{\theta})\,\right)}_{\text{Sec.~\ref{embedding_classifier}}}\underbrace{p(c\,|\,\mathcal{Y}', \Vec{\theta})}_{\text{Sec.~\ref{extracting_emb}}}\Big].
\end{align}
which indicates the log-likelihood of the additional labels are weighted by the grouping of their semantic relationships. We use the $f_{sc}$ in the paper to denote the classifier head that predicts these additional labels. 

\section{Obtaining Label Embeddings using ConceptNet Knowledge Graph}
In this section, we elaborate on the process described in Sec.~\ref{extracting_emb} which aims to obtain class label embeddings which correlate well with visual similarity. We start by describing the procedure and then we present some qualitative examples to demonstrate its effectiveness. 

\subsection{Procedure}
We follow a similar procedure to that described in~\cite{speer2016_conceptnet} with one crucial difference. Instead of using the entire ConceptNet graph, we use the graph after filtering it to retain only the relations which imply visual similarity (\emph{see} Sec.~\ref{experiments} for more details). 

We start with the filtered graph, the GloVe word embeddings matrix, and the word2vec word embeddings matrix. The process comprises two main steps: 1) retrofitting each of the GloVe and word2vec embeddings using the ConceptNet filtered graph to obtain two new sets of embeddings, and 2) combining the two retrofitted sets to obtain our final hybrid embeddings set.

\smallskip
\noindent \textbf{Retrofitting}
Given the filtered graph and a matrix of word embeddings, the aim is to infer for each term/word a new embedding vector $\Vec{v}_i$ which is close to the original vector $\hat{\Vec{v}}_i$ but also close to the term neighbors in the graph with edges $E$. This can be achieved by minimizing the following objective function.
\begin{align}
E(v) = \sum_{i=1}^{n} \left[\alpha_i\norm{\Vec{v}_i - \hat{\Vec{v}}_i}^2 + \sum_{(i,j) \in E} \beta_{ij}\norm{\Vec{v}_i - \Vec{v}_j}^2 \right] ;\nonumber
\end{align}
with $\alpha_i=1$ if term $i$ is present in the embeddings vocabulary and zero otherwise; and $\beta_{ij}$ denoting the weight of the edge connecting term $i$ and term $j$. Note that the use of $\alpha$ allows optimizing the above objective for terms which appear in the knowledge graph even if it is not present in the vocabulary of the word embeddings~\cite{speer2016_ensemble_retro}. 
To minimize the above function, we follow the iterative algorithm originally suggested by Faruqui \etal~\cite{faruqui2014_retrofitting} and later extended by Speer \etal~\cite{speer2016_ensemble_retro}. We perform such optimization twice: once for the GloVe embeddings and another for the word2vec.

\smallskip
\noindent \textbf{Combining the Two Sets}
After applying retrofitting to both matrices, we combine them by finding a globally linear projection that aligns the results based on their common vocabulary. As inspired by~\cite{zhao2015_learning_trans} and \cite{speer2016_conceptnet}, to find such projection, we concatenate the columns of the two matrices and use SVD to reduce their dimensionality to 128. Such alignment allows us to deduce compatible embeddings for terms which appear in one of the vocabularies but not the other. This alignment and merging give rise to a hybrid set of embeddings which combines all three sources: GloVe, word2vec, and ConceptNet filtered graph. We use this set as the basis for establishing the prior on visual similarity among a given set of class labels (\emph{see} Sec.~\ref{sec:our_method}). 
\begin{figure*}[!]
    \centering
  \includegraphics[width=0.99\textwidth]{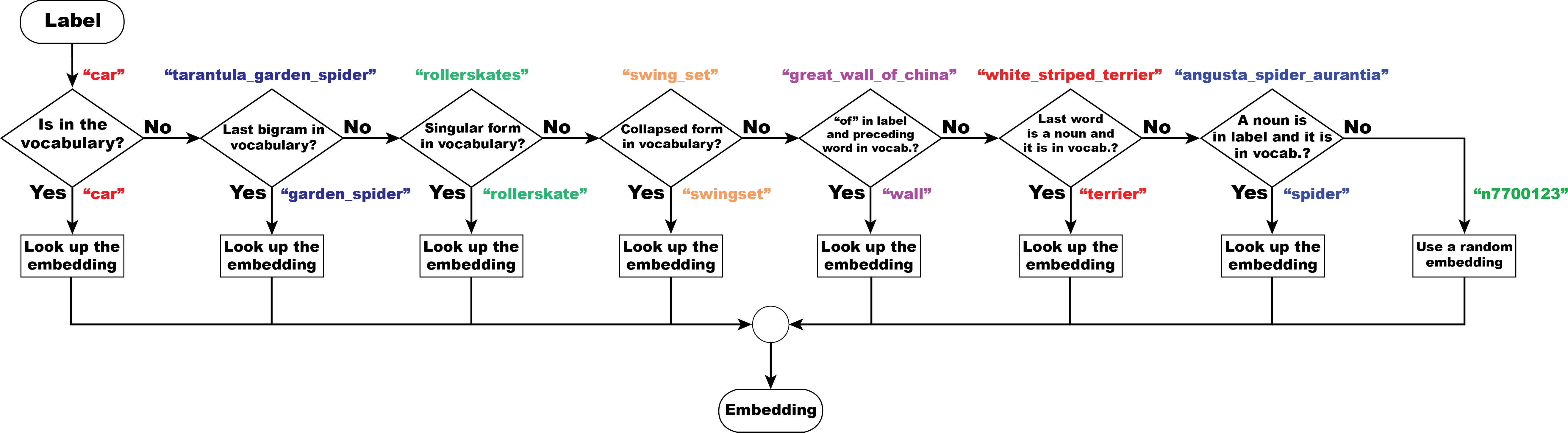}
 \caption{A flowchart describing our label embedding lookup strategy aiming to find the most reasonable embedding for a given class label. We include demonstrative examples for each of the fall-out cases.}
 \label{fig:vocab_fallout}
\end{figure*}

\smallskip
\noindent \textbf{Handling Out-of-Vocabulary Labels}
Our obtained embeddings vocabulary consists of approximately 500k different terms and hence provides a reasonable coverage for most of the class labels. However, it might sometimes be the case that one or more of the class labels are missing from the vocabulary. In such event, we employ a fall-out strategy to find the most reasonable alternative. We present the flowchart for our fall-out strategy in Fig.~\ref{fig:vocab_fallout}.

\subsection{Label Grouping Examples}
As mentioned in Sec.~\ref{sec:our_method}, we apply density-based clustering on the class labels embeddings to group the labels into visually similar concepts. To demonstrate the effectivness of the retrofitted embeddings in capturing said similarity, we compare the clustering output if we use the retrofitted embeddings as opposed to if we use the GloVe distributional embeddings without retrofitting. We perform this comparison for Mini-ImageNet (\emph{see} Table~\ref{mini_retro},\ref{mini_glove}), CIFAR-100 (\emph{see} Table~\ref{cifar_retro},\ref{cifar_glove}), and DomainNet classes (\emph{see} Table~\ref{domain_retro},\ref{domain_glove}). Note that we only report the groups having more than one member and we omit single-member groups. We observe that in all three cases, clustering the retrofitted embeddings produces groups which largely match our intuition about visual similarity. On the other hand, we notice that clustering the non-retrofitted GloVe embeddings results in grouping labels which usually appear in similar context, even if they are not visually similar. For example, in Table~\ref{cifar_glove}, we observe that \emph{``sea''} was grouped with other classes which are contextually related to \emph{``sea''}, yet bear no visual similarity to it. This is due to the fact that GloVe embeddings are learnt in a way that captures distributional semantics rather than visual semantics. However, when the GloVe embeddings are retrofitted with the ConceptNet filtered graph, we witness an improved grouping which aligns better with visual semantic similarity.

\section{Implementation Details}
\begin{figure}[h!]
\centering
  \includegraphics[width=6cm,scale=0.8]{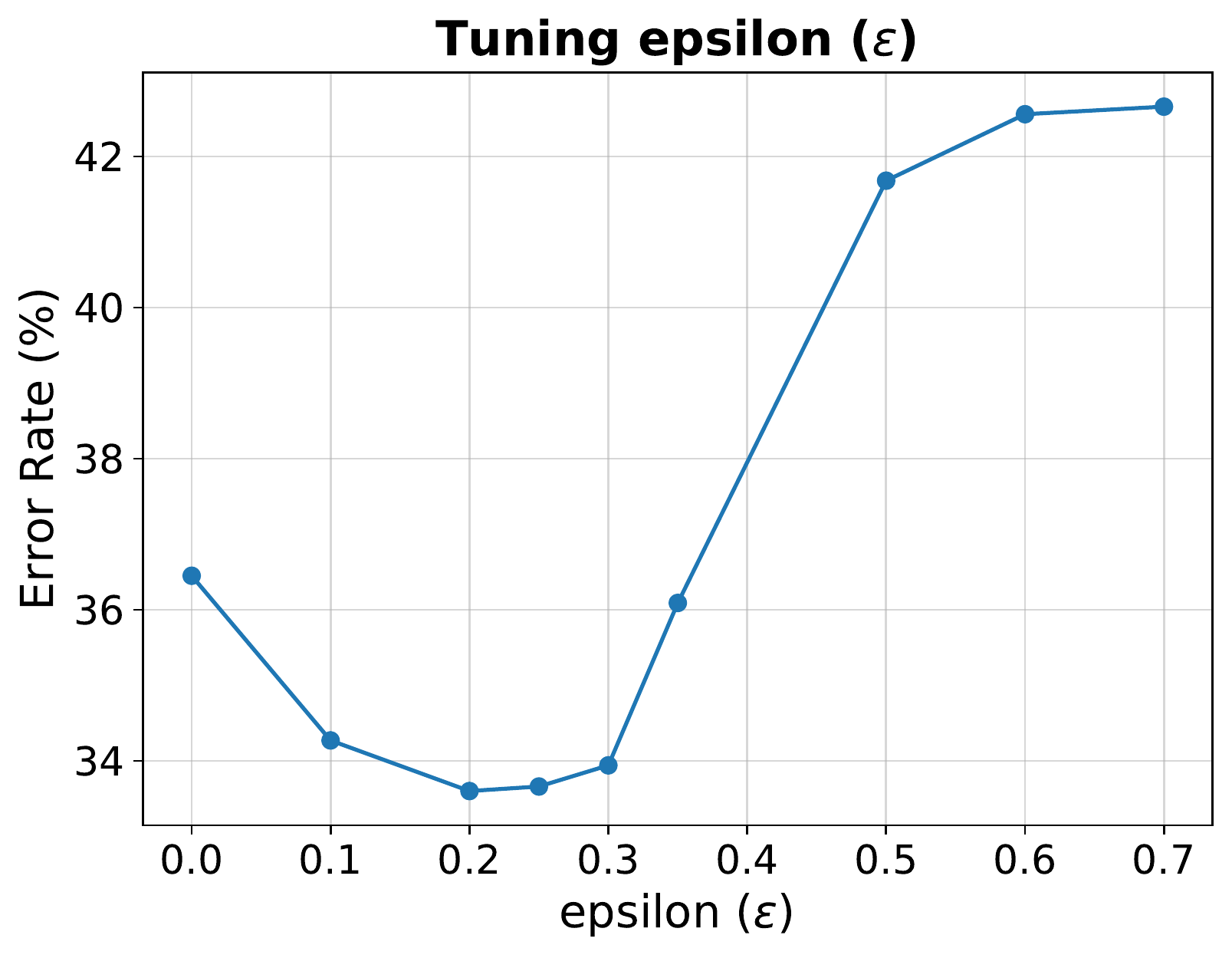}
 \caption{Error rates for different values of $\epsilon$. $\epsilon=0$ corresponds to no label grouping, while $\epsilon > 0.7$ corresponds to grouping all labels into a single cluster.}
 \label{fig:tuning_epsilon}
\end{figure}

\smallskip
\noindent \textbf{Hyperparameters}
In our preliminary experiments, we mostly found that our method is not sensitive to the hyperparameters, so we tuned their values via a validation set on a single experiment (CIFAR100 - 2500 labels) then fixed them across all other experiments to the values shown in Table~\ref{hyperparams_table}. The only exception is the density-based clustering parameter $\epsilon$. The number of clusters (i.e. label groups) is automatically decided based on $\epsilon$, which denotes the maximum cosine distance between two embedding vectors for one to be considered in the same neighbourhood as the other. The larger the value of $\epsilon$, the more aggressive the grouping is (\ie the more members in each group). Accordingly, we tune $\epsilon$ individually for each dataset. We find that $\epsilon=0.2$ works well for all datasets except Mini-ImageNet where we use $\epsilon=0.3$ instead. In Fig.~\ref{fig:tuning_epsilon}, we demonstrate the effect of varying $\epsilon$ on the error rate using a single split of CIFAR-100 (2500 labeled instances) when training for 100 epochs.

\begin{table*}[t]
\centering
\small
\caption{Hyper-parameters values across all our experiments}
\label{hyperparams_table}
\begin{tabular}{llllc}
\hline
\multicolumn{1}{c}{\textbf{Hyper-parameter}} &
   &
  \multicolumn{1}{c}{\textbf{Description}} &
   &
  \textbf{Value} \\ \cline{1-1} \cline{3-3} \cline{5-5} 
$\lambda_u$        &  & Unlabeled loss coefficient                                         &  & $1.00$        \\
$\lambda_{co}$     &  & Co-training loss coefficient                                       &  & $1.00$        \\
$\tau_e$           &  & Pseudo-labeling confidence threshold for the \emph{Semantic Classifier}       &  & $0.70$      \\
$\tau_o$           &  & Pseudo-labeling confidence threshold for the \emph{One-Hot Classifier}        &  & $0.95$     \\
$batch\_size$         &  & Number of labeled images per batch                                 &  & $64$       \\
$\mu$               &  & Ratio between number of unlabeled and labeled images in each batch &  & $3$        \\
$images\_per\_epoch$ &  & Number of labeled images per epoch                                 &  & $64\times1024$  \\
$num\_epochs$        &  & Number of epochs of training                                       &  & $300$      \\
$lr$      &  & learning rate max value (10 epochs warmup then cosine decay)       &  & $0.03$     \\
${weight}\_{decay}$      &  & Weight decay regualrization coefficient                            &  & $5.00\times10^{-4}$ \\
$momentum$           &  & Nesterov momentum for SGD optimizer                                &  & $0.90$      \\
$emb\_dim$           &  & Dimensionality of the label embeddings                             &  & $128$      \\
$\epsilon$ &
   &
  DBSCAN clustering coefficient   denoting the maximum distance between \\ 
  && two samples for one to be considered as   in the neighborhood of the other &
   &
  $0.20$ \\ \hline
\end{tabular}%
\end{table*}

\smallskip
\noindent \textbf{Semantic Classifier Loss}
We use two different loss functions for our two classifiers, \ie \emph{cosine loss} for the \emph{Semantic Classifier}, and \emph{cross-entropy} for the \emph{One-Hot Classifier} (\emph{see} Sec.~\ref{sec:our_method}). It is, hence, important to consider the scale of both losses so that one doesn't dominate over the other. \emph{Cosine loss} values are bounded between 0 and 2 while \emph{cross-entropy} values are not. Accordingly, we multiply the \emph{Semantic Classifier} loss by a factor of 3 before applying the back propagation step. We obtained such value by using a held-out validation set on CIFAR-100 (1000 labeled examples) and we fixed it across all other experiments and datasets. 

\smallskip
\noindent \textbf{Augmentations}
As described in Sec.~\ref{sec:background}, we make use of two types of augmentations, \ie weak and strong. For weak augmentations, we use random cropping and padding, and random horizontal flips. As for the strong augmentations, we use the RandAugment~\cite{cubuk2020_randaugment} list of transformations for both our system and the FixMatch baseline. 

\begin{figure*}[t!]
    \centering
  \includegraphics[width=0.99\textwidth]{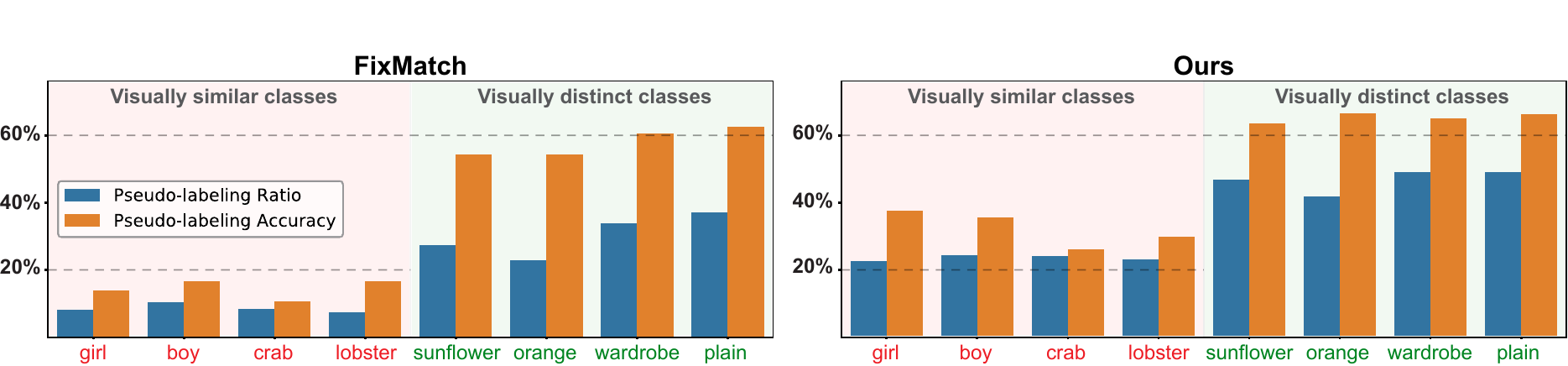}
 \caption{CIFAR-100 confidence-based pseudo-labeling comparison between the baseline (left) and our method (right). \emph{Accuracy} values show how much, on average, pseudo-labels for a given class match the true label, while \emph{Ratio} values show the percentage of samples of a given class which are retained for pseudo-labeling (i.e. with confidence score above the threshold). The two metrics are calculated for the 4 most (red) and least (green) visually similar classes over the first 10 epochs of training.}
 \label{fig:pl_stats_comparison_cifar}
\end{figure*}

\begin{figure*}[h!]
\centering
  \begin{subfigure}[b]{0.45\linewidth}
    \includegraphics[width=6.5cm, scale=0.6]{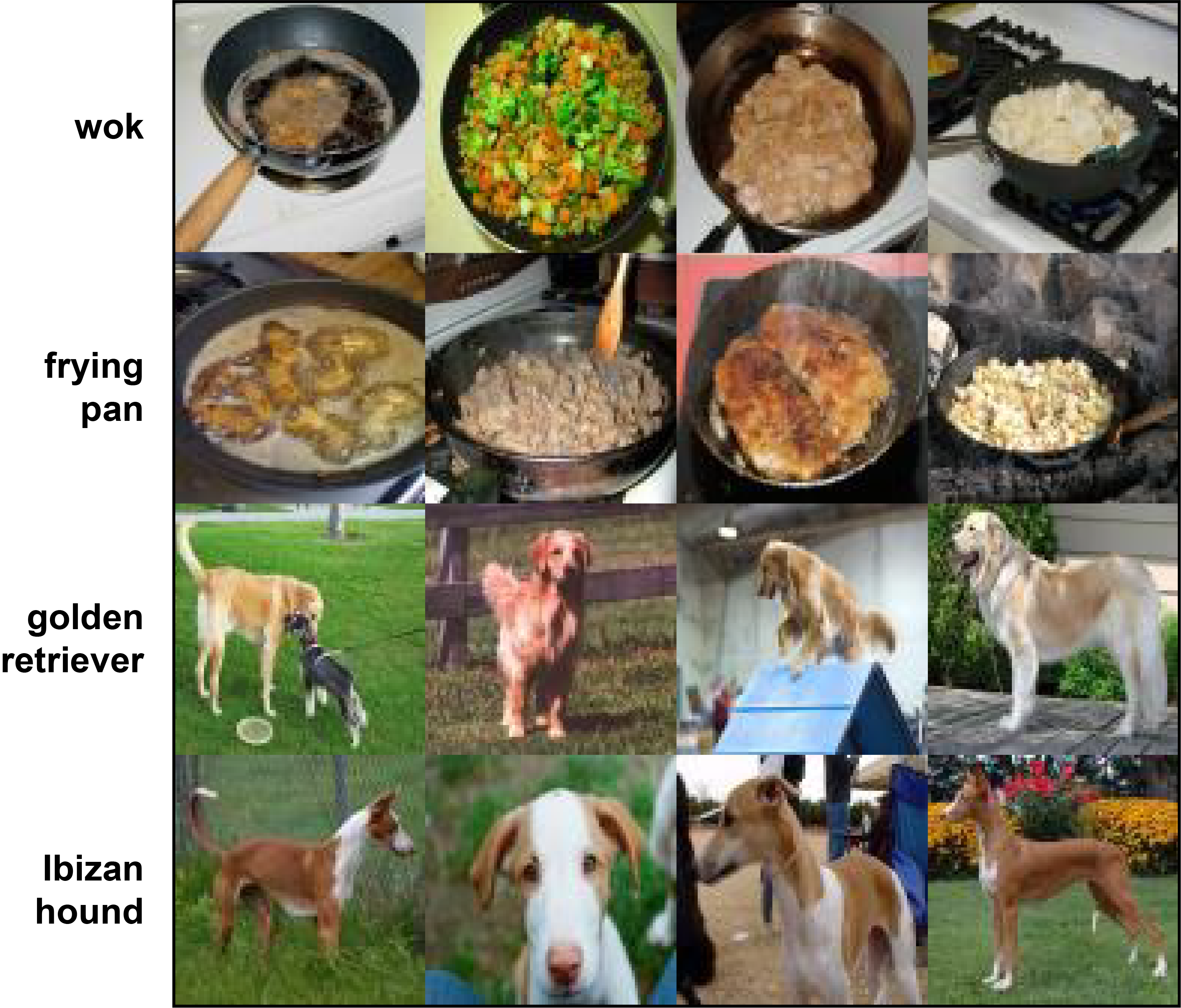}
    \caption{Mini-ImageNet}
  \end{subfigure}
  \begin{subfigure}[b]{0.45\linewidth}
    \includegraphics[width=6.5cm, scale=0.6]{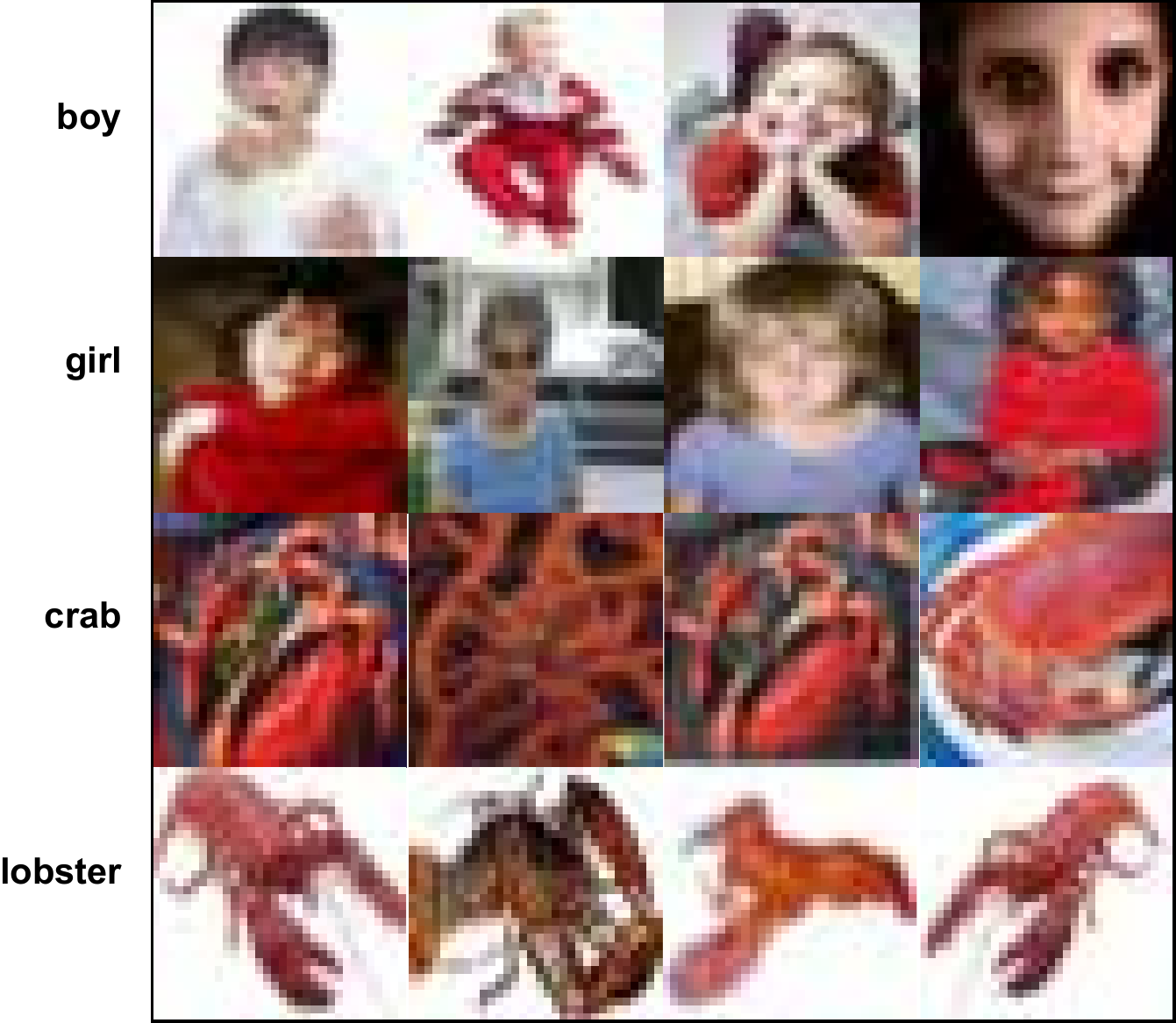}
    \caption{CIFAR-100}
  \end{subfigure}
  \caption{The most confused images for the 4 most visually similar classes of Mini-ImageNet (left) and CIFAR-100 (right). The caption next to each image group denotes the true class to which the image group belongs.}
  \label{fig:confusing_images_2in1}
\end{figure*}

\smallskip
\noindent \textbf{Inference}
Since we train two classifiers in our method, during inference time, we can choose one of three options for inference: 1) use the \emph{One-Hot Classifier} prediction, 2) use the \emph{Semantic Classifier} prediction, 3) Average the softmax scores of the two classifiers and use the combined score for prediction. During our validations, we found that the former approach always yields marginally better results, so we use it as our basis for inference. Finally, We also use an exponential moving average of model weights with a decay rate of 0.999.

\section{Further Pseudo-labeling Analysis}
In Fig.~\ref{fig:pl_stats_comparison} in the main text, we present a comparison between pseudo-labeling statistics (on Mini-ImageNet dataset) of our method versus the baseline. In this section, we elaborate about the experimental setup for obtaining these statistics. Additionally, we provide similar analysis on CIFAR-100 dataset. 

For a given dataset, we run our algorithm for 10 epochs of unlabeled data and we capture a highly granular view of the pseudo-labeling statistics for each mini-batch. Consequently, we calculate two metrics: 1) we use the true labels of the unlabeled data samples (which we originally ignore to emulate an SSL setting) to measure the true pseudo-labeling accuracy for each of the classes in the dataset, and 2) we use the classifier confidence scores to calculate the pseudo-labeling ratio for each class, which represents the amount of unlabeled samples exceeding the confidence threshold and thereby are retained for pseudo-labeling. We repeat the same procedure and measure the same metrics for our baseline \cite{Sohn_fixmatch20}. We, then, display those metrics for the 4 classes which were deemed by our clustering method as the most visually similar concepts. Conversely, we also display them for the 4 classes which are deemed most visually distinct.   In Fig.~\ref{fig:pl_stats_comparison_cifar}, we report these metrics for CIFAR-100 dataset (\emph{see} Fig.~\ref{fig:pl_stats_comparison} for Mini-ImageNet).
Additionally, through the same experimental setup described above, we keep track of pseudo-labeling statistics for each individual unlabeled image. We report in Fig.~\ref{fig:confusing_images_2in1} the most confused images among the 4 most visually similar classes for both datasets. We define confusion as the average number of times a given image is incorrectly pseudo-labeled as another class within the 4 classes (\eg \emph{``boy''} pseudo-labeled as \emph{``girl''}).

\begin{table*}[h]
\centering
\small
\caption{Mini-ImageNet class groups obtained by clustering the retrofitted embeddings.}
\label{mini_retro}
\begin{tabular}{lll}
\hline
Group      &  & Members                             \\ \hline
Group    1 &  & 'French\_bulldog', 'Ibizan\_hound', 'Saluki', 'Walker\_hound',   'golden\_retriever', 'malamute', 'miniature\_poodle' \\
Group    2 &  & 'catamaran', 'yawl'                 \\
Group    3 &  & 'frying\_pan', 'wok'                \\
Group    4 &  & 'horizontal\_bar', 'parallel\_bars' \\ \hline
\end{tabular}%
\end{table*}


\begin{table*}[h]
\small
\centering
\caption{Mini-ImageNet class groups obtained by clustering the GloVe embeddings.}
\label{mini_glove}
\begin{tabular}{lll}
\hline
Group      &  & Members                                                                                                                  \\ \hline
Group    1 &  & 'African\_hunting\_dog', 'French\_bulldog', 'Ibizan\_hound', 'Walker\_hound',   'golden\_retriever', 'miniature\_poodle' \\
Group    2 &  & 'combination\_lock', 'garbage\_truck', 'horizontal\_bar', 'parallel\_bars',   'pencil\_box', 'street\_sign'              \\ \hline
\end{tabular}
\end{table*}


\begin{table*}[!]
\centering
\caption{CIFAR-100 class groups obtained by clustering the retrofitted embeddings.}
\label{cifar_retro}
\begin{tabular}{lll}
\hline
Group      &  & Members                               \\ \hline
Group    1 &  & 'aquarium\_fish', 'flatfish', 'trout' \\
Group    2 &  & 'bicycle', 'motorcycle'               \\
Group    3 &  & 'boy', 'girl'                         \\
Group    4 &  & 'crab', 'lobster'                     \\
Group    5 &  & 'dolphin', 'whale'                    \\
Group    6 &  & 'man', 'woman'                        \\
Group    7 &  & 'oak\_tree', 'pine\_tree'             \\ \hline
\end{tabular}%
\end{table*}


\begin{table*}[t]
\small
\centering
\caption{CIFAR-100 class groups obtained by clustering the GloVe embeddings.}
\label{cifar_glove}
\begin{tabular}{lll}
\hline
Group      &  & Members                        \\ \hline
Group    1 &  & 'aquarium\_fish', 'crab', 'dolphin', 'lobster', 'sea', 'shark', 'trout',   'turtle', 'whale'                                   \\
Group    2 &  & 'elephant', 'fox',  'house', 'leopard', 'lion', 'man',   'pickup\_truck', 'road', 'table', 'tiger', 'tractor', 'wolf', 'woman' \\
Group    3 &  & 'bicycle', 'motorcycle'        \\
Group    4 &  & 'bus', 'train'                 \\
Group    5 &  & 'crocodile', 'lizard', 'snake' \\
Group    6 &  & 'raccoon', 'squirrel'          \\
Group    7 &  & 'oak\_tree', 'pine\_tree'      \\ \hline
\end{tabular}%
\end{table*}


\begin{table*}[t]
\centering
\caption{DomainNet class groups obtained by clustering the retrofitted embeddings.}
\label{domain_retro}
\begin{tabular}{lll}
\hline
Group      &  & Members                        \\ \hline
Group    1 &  & 'basketball', 'soccer\_ball'   \\
Group    2 &  & 'beard', 'goatee', 'moustache' \\
Group    3 &  & 'bicycle', 'motorbike'         \\
Group    4 &  & 'birthday\_cake', 'cake'       \\
Group    5 &  & 'bracelet', 'necklace'         \\
Group    6 &  & 'cello', 'clarinet', 'guitar', 'piano', 'saxophone', 'trombone',   'trumpet', 'violin' \\
Group    7 &  & 'crab', 'lobster'              \\
Group    8 &  & 'oven', 'stove'                \\
Group    9 &  & 'pants', 'shorts', 'underwear' \\
Group 10   &  & 'pickup\_truck', 'truck'       \\
Group 11   &  & 'wine\_bottle', 'wine\_glass'  \\ \hline
\end{tabular}%
\end{table*}

\begin{table*}[h!]
\vspace{-8cm}
\centering
\caption{DomainNet class groups obtained by clustering the GloVe embeddings.}
\label{domain_glove}
\begin{tabular}{lll}
\hline
Group       &  & Members                                                           \\ \hline
Group    1  &  & 'airplane', 'helicopter'                                          \\
Group    2  &  & 'ambulance', 'hospital'                                           \\
Group    3  &  & 'apple', 'blackberry'                                             \\
Group    4  &  & 'asparagus', 'broccoli', 'peas'                                   \\
Group    5  &  & 'axe', 'knife', 'sword'                                           \\
Group    6  &  & 'backpack', 'suitcase'                                            \\
Group    7  &  & 'banana', 'blueberry', 'pineapple', 'strawberry'                  \\
Group    8  &  & 'baseball', 'basketball'                                          \\
Group    9  &  & 'baseball\_bat', 'bat'                                            \\
Group    10 &  & 'bathtub', 'sink', 'toilet'                                       \\
Group    11 &  & 'beard', 'goatee', 'moustache'                                    \\
Group    12 &  & 'bracelet', 'necklace'                                            \\
Group    13 &  & 'bread', 'cake', 'cookie', 'peanut', 'pizza', 'sandwich'          \\
Group    14 &  & 'bus', 'train'                                                    \\
Group    15 &  & 'carrot', 'onion', 'potato'                                       \\
Group    16 &  & 'crab', 'dolphin', 'fish', 'lobster', 'octopus', 'shark', 'whale' \\
Group    17 &  & 'crayon', 'pencil'                                                \\
Group    18 &  & 'fireplace', 'microwave', 'oven', 'stove'                         \\
Group    19 &  & 'jacket', 'pants', 'shorts', 'sweater', 'underwear'               \\
Group    20 &  & 'raccoon', 'squirrel'                                             \\
Group    21 &  & 'radio', 'television'                                             \\
Group    22 &  & 'snowflake', 'snowman'                                            \\
Group    23 &  & 'toothbrush', 'toothpaste'                                        \\ \hline
\end{tabular}%
\end{table*}